\documentclass[11pt, letterpaper]{article}

\usepackage{amsmath}
\usepackage{amssymb}
\usepackage{amsthm}
\usepackage{graphicx}
\usepackage{hyperref}
\usepackage{url}
\usepackage{algorithm}
\usepackage{algorithmic}
\usepackage{caption}
\usepackage{subcaption}
\usepackage{booktabs}
\usepackage{authblk}
\usepackage{appendix}
\usepackage[utf8]{inputenc}
\usepackage[T1]{fontenc}
\usepackage[english]{babel}
\usepackage{geometry}
\usepackage{xcolor}
\usepackage{fancyhdr}
\usepackage{microtype}
\usepackage{parskip}
\usepackage{tabularx}
\usepackage{enumitem}
\usepackage{multirow}
\usepackage{booktabs}
\usepackage{adjustbox}
\usepackage{multicol}
\usepackage{CJKutf8}
\usepackage[labelfont=bf, justification=justified, singlelinecheck=false]{caption}

\hypersetup{
    colorlinks=true,
    linkcolor=blue,
    filecolor=magenta,      
    urlcolor=cyan,
    citecolor=red,
}

\definecolor{headercolor}{RGB}{0, 50, 100}
\title{RadLLM: A Comprehensive Healthcare Benchmark of Large Language Models for Radiology}

\date{}

\newcommand*\samethanks[1][\value{footnote}]{\footnotemark[#1]}
\author[1]{Zhengliang Liu \thanks{Co-first authors.}}
\author[2]{Tianyang Zhong \samethanks}
\author[1]{Yiwei Li \samethanks}
\author[8]{Yutong Zhang \thanks{Co-second authors.}}
\author[5,6,1]{Yi Pan \samethanks}
\author[11]{Zihao Zhao \samethanks}
\author[2]{Peixin Dong \samethanks}
\author[4]{Chao Cao \samethanks}
\author[11,20]{Yuxiao Liu \samethanks}
\author[1]{Peng Shu \samethanks}
\author[2]{Yaonai Wei \samethanks}
\author[1]{Zihao Wu}
\author[2]{Chong Ma}
\author[10]{Jiaqi Wang}
\author[17]{Sheng Wang}
\author[2]{Mengyue Zhou}
\author[2]{Zuowei Jiang}
\author[2]{Chunlin Li}
\author[7]{Jason Holmes}
\author[1]{Shaochen Xu}
\author[4]{Lu Zhang}
\author[1]{Haixing Dai}
\author[18]{Kai Zhang}
\author[1]{Lin Zhao}
\author[19]{Yuanhao Chen}
\author[2]{Xu Liu}
\author[7]{Peilong Wang}
\author[1]{Junhao Chen}

\author[14]{Pingkun Yan}
\author[15]{Jun Liu}
\author[9]{Bao Ge}
\author[18]{Lichao Sun}
\author[4]{Dajiang Zhu}
\author[3]{Xiang Li}
\author[7]{Wei Liu}
\author[2]{Xiaoyan Cai}
\author[2]{Xintao Hu}
\author[5]{Xi Jiang}
\author[10]{Shu Zhang}
\author[8]{Xin Zhang}
\author[2]{Tuo Zhang}
\author[2]{Shijie Zhao}
\author[3]{Quanzheng Li}
\author[13]{Hongtu Zhu}
\author[11,12,16]{Dinggang Shen}
\author[1]{Tianming Liu \thanks{Corresponding author. Email: tianming.liu@gmail.com.}}

\affil[1]{School of Biomedical Engineering, ShanghaiTech University, and Shanghai Clinical Research and Trial Center, Shanghai 201210, ChinaComputing, University of Georgia, GA, USA}
\affil[2]{School of Automation, Northwestern Polytechnical University, Xi'an 710072, China}
\affil[3]{Department of Radiology, Massachusetts General Hospital and Harvard Medical School, MA, USA}
\affil[4]{Department of Computer Science and Engineering, University of Texas at Arlington, TX, USA}
\affil[5]{School of Life Science and Technology, University of Electronic Science and Technology of China, Chengdu 611731, China}
\affil[6]{Glasgow College, University of Electronic Science and Technology of China, Chengdu 611731, China }
\affil[7]{Department of Radiation Oncology, Mayo Clinic, Phoenix, Arizona, USA}
\affil[8]{Institute of Medical Research, Northwestern Polytechnical University, Xi'an 710072, China}
\affil[9]{School of Physics and Information Technology, Shaanxi Normal University, Xi’an 710119 China}
\affil[10]{School of Computer Science, Northwestern Polytechnical University, Xi'an 710072, China}
\affil[11]{School of Biomedical Engineering, ShanghaiTech University, and Shanghai Clinical Research and Trial Center, Shanghai 201210, China}
\affil[12]{Shanghai United Imaging Intelligence Co., Ltd.}
\affil[13]{Department of Biostatistics, University of North Carolina at Chapel Hill, Chapel Hill, NC 27599, USA}
\affil[14]{Department of Biomedical Engineering and Center for Biotechnology and Interdisciplinary Studies at Rensselaer Polytechnic Institute, Troy, New York 12180, USA}
\affil[15]{Department of Radiology, Second Xiangya Hospital, Changsha 410011, China}
\affil[16]{Shanghai Clinical Research and Trial Center}
\affil[17]{School of Biomedical Engineering, Shanghai Jiao Tong University, Shanghai 200240, China}
\affil[18]{Department of Computer Science and Engineering, Lehigh University, PA, USA}
\affil[19]{Department of Linguistics and Department of Computer Science, Dartmouth College, NH, USA}
\affil[20]{Lingang Laboratory, Shanghai, 200031, China}

\begin{document}

\maketitle

\begin{abstract}
The rise of large language models (LLMs) has marked a pivotal shift in the field of natural language processing (NLP). LLMs have revolutionized a multitude of domains, and they have made a significant impact in the medical field. Large language models are now more abundant than ever, and many of these models exhibit bilingual capabilities, proficient in both English and Chinese. However, a comprehensive evaluation of these models remains to be conducted. This lack of assessment is especially apparent within the context of radiology NLP. This study seeks to bridge this gap by critically evaluating thirty two LLMs in interpreting radiology reports, a crucial component of radiology NLP. Specifically, the ability to derive impressions from radiologic findings is assessed. The outcomes of this evaluation provide key insights into the performance, strengths, and weaknesses of these LLMs, informing their practical applications within the medical domain.

\end{abstract}

\section{Introduction}
In recent years, large language models (LLMs) \cite{liu2023summary,zhou2023comprehensive,holmes2023evaluating,wu2023exploring,li2023artificial,rezayi2023exploring,liu2022survey,zhao2023brain} have emerged as prominent tools in the realm of natural language processing (NLP). Compared with traditional NLP models, LLMs are trained on expansive datasets and demonstrate impressive capabilities ranging from language translation to creative content generation, and problem-solving.  For example, OpenAI's dialogue model, ChatGPT\footnote{\url{https://openai.com/blog/chatgpt}}, has garnered widespread attention due to its outstanding performance, sparking a trend in the development of LLMs that has had profound effects on the growth of the entire AI community.

ChatGPT was developed based on GPT-3.5 version \cite{brown2020language}, released in November 2022. It is trained on a massive dataset of text and code and can generate text, translate languages, write different kinds of creative content, and answer your questions in an informative way. It is currently widely used in many areas, such as intelligent customer service, summary generation, and more, offering greater possibilities for applications of language models \cite{lund2023chatting,sallam2023chatgpt,liu2023summary,sallam2023utility}. The multimodal model GPT-4\footnote{\url{https://openai.com/research/gpt-4}} with 1.8 trillion parameters was released in March 2023, with overall performance and accuracy surpassing the previous version. Also, the introduction of ChatGPT plugin functionality has directly endowed ChatGPT with the ability to use other tools and connect to the internet, breaking the constraints of the model's data. The popular rise of ChatGPT has led to a surge in the development of new LLMs. There are now several hundred open-source LLMs available, such as Hugging Face's BLOOM \cite{scao2022bloom}, which is trained with 176 billion parameters across 46 natural languages and 13 programming languages while most LLMs are not publicly released and are mainly based on Latin languages with English as the main language. Also, starting from Meta's open source LLaMA \cite{touvron2023llama} series of models, researchers from Stanford University and other institutions have successively open-sourced LLaMA-based lightweight classes such as Alpaca\footnote{\url{https://crfm.stanford.edu/2023/03/13/alpaca.html}}, Koala\footnote{\url{https://bair.berkeley.edu/blog/2023/04/03/koala/}}, and Vicuna\footnote{\url{ https://github.com/lm-sys/FastChat}}, etc. emerged. The research and application threshold of this type of model is greatly reduced, and the training, and reasoning costs have been repeatedly reduced.

Similarly, LLM ecology in other countries such as China has also begun to take shape. At present, the LLMs in China can basically be divided into three tracks: companies, institutions, and universities, such as Baidu's ERNIE Bot\footnote{\url{https://yiyan.baidu.com/}}, Huawei's Pangu models series, IDEA's Ziya-LLaMA\footnote{\url{https://github.com/IDEA-CCNL/Fengshenbang-LM}} series, Fudan University's MOSS\footnote{\url{https://github.com/OpenLMLab/MOSS}}. Many corresponding models are based on LLaMA, chatGLM\footnote{\url{https://github.com/THUDM/ChatGLM-6B}}, BLOOM, and other models that do not follow the transformer approach, such as baichuan\footnote{\url{https://modelscope.cn/models/baichuan-inc/Baichuan-13B-Base/summary}}, RWKV \cite{peng2023rwkv}. These models are being used for various tasks, including natural language understanding, natural language generation, machine translation, and question answering.

The versatility of these models extends into the medical field \cite{eggmann2023implications,lievin2022can,wang2023large,taylor2022galactica,singhal2023towards,ufuk2023role,yunxiang2023chatdoctor,haupt2023ai,zhao2023brain}. For example, LLMs can be used to generate personalized medical reports \cite{liao2023differentiate,dai2023chataug,liu2023deid,ma2023impressiongpt}, facilitate online medical consultations, remote medical diagnosis, and guidance \cite{li2023artificial,rezayi2022clinicalradiobert}, and aid in medical data mining \cite{li2023artificial}, among other applications. In radiology and medical image analysis, \cite{pons2016natural,casey2021systematic,adams2023leveraging,doshi2023utilizing,lyu2023translating,ali2023using,balagopal2021psa,liu2023radiology,zhang2023segment,zhang2023differentiating,zhang2023beam}, which are fields that have been continuously intertwined with developments in AI \cite{liu2023radiology,zhou2023fine,dai2023samaug,bi2023community,zhang2023segment,ding2023deep,ding2022accurate,qiang4309357deep,dai2022graph}, one significant application lies in the interpretation of images. Generative AI can help automate the process of preliminary diagnosis, potentially saving physicians' time. It could be particularly useful in situations where there is a shortage of trained radiologists. Moreover, physicians may no longer need to manually enter data into the patient's electronic medical record. In addition, these models are capable of aiding in clinical decision-making. By processing and analyzing a patient's radiological data along with other relevant medical information, LLMs can generate patient-specific reports and provide possible diagnoses, treatment recommendations, or potential risks. Furthermore, advances in LLM have led to the development of many specialized biomedical LLMs, such as HuatuoGPT 
\cite{zhang2023huatuogpt}, an open-source model from the Chinese University of Hong Kong with the BLOOMZ \cite{muennighoff2023crosslingual} as backbone, uses the data distilled from ChatGPT and the real data of doctors in the supervised fine-tuning stage. And XrayGLM,\footnote{\url{https://github.com/WangRongsheng/XrayGLM}}, the first Chinese multimodal large model dedicated to the diagnosis of chest X-rays, which is based on VisualGLM-6B\footnote{\url{https://github.com/THUDM/VisualGLM-6B}} and then fine-tuned on two open chest X-rays datasets. Others include QiZhenGPT\footnote{\url{https://github.com/CMKRG/QiZhenGPT}}, BioMedLM\footnote{\url{http://github.com/standford-crfm/BioMedLM.html}}, BioGPT \cite{luo2022biogpt}, PMC-LLaMA \cite{wu2023pmc}, Med-PaLM \cite{singhal2023large}, etc., which demonstrate significant potential of LLMs in the medical field. 

Despite the escalating ubiquity of LLMs in various sectors, a comprehensive understanding and evaluation of their performance, particularly in the specialized field of radiology NLP, remains noticeably absent. This paucity of knowledge is even more stark when we consider the emerging LLMs developed in other countries such as China, a significant portion of which boast robust bilingual capabilities in both English and Chinese. Often untapped and under-evaluated, these models could offer unique advantages in processing and understanding multilingual medical data. The scarcity of in-depth, scientific performance evaluation studies on these models in the medical and radiology domains signals a significant knowledge gap that needs addressing. Given this backdrop, we believe it is paramount to undertake a rigorous and systematic exploration and analysis of these world-wide LLMs. This would not only provide a better understanding of their capabilities and limitations but also position them within the global landscape of LLMs. By comparing them with established international contenders, we aim to shed light on their relative strengths and weaknesses, providing a more nuanced understanding of the application of LLMs in the field of radiology. This, in turn, would potentially contribute to the optimization and development of more efficient and effective NLP/LLM tools for radiology.

Our study is focused on the crucial aspect of radiology NLP, namely, interpreting radiology reports and deriving impressions from radiologic findings. We rigorously evaluate the selected models using a robust dataset of radiology reports, benchmarking their performance against a variety of metrics.

Initial findings reveal that distinct differences were observed in the models' respective strengths and weaknesses. The implications of these findings, as well as their potential impact on the application of LLMs in radiology NLP, are discussed in detail.

In the grand scheme, this study serves as a pivotal step towards the wider adoption and fine-tuning of LLMs in radiology NLP. Our observations and conclusions are aimed at spurring further research, as we firmly believe that these LLMs can be harnessed as invaluable tools for radiologists and the broader medical community.
\section{Related work}

\subsection{Evaluating Large Language Models}
The rapid development of LLMs has been revolutionizing the field of natural language processing \cite{J1,J2,J3} and domains that benefit from NLP \cite{rezayi2022agribert,liu2023context,dai2023ad,cai2022coarse,liu2023radiology,liao2023mask,cai2023exploring,rezayi2022clinicalradiobert,zhao2022embedding,zhao2023generic}. These powerful models have shown significant performance in many NLP tasks, like natural language generation (NLG) and even Artificial General Intelligence (AGI). However, utilizing these models effectively and efficiently requires a practical understanding of their capabilities and limitations, and overall performance, so evaluating these models is of paramount importance. 

To compare the capabilities of different LLMs, researchers usually test with benchmark datasets in various fields (such as literature, chemistry, biology, etc.), and then evaluate their performance according to traditional indicators (such as correct answer rate, recall rate, and F1 value). The most recent study from OpenAI \cite{J4} includes the pioneering research study that assesses the performance of large language models (i.e. GPT-4) on academic and professional exams specifically crafted for educated individuals. The findings demonstrate exceptional performance of GPT-4 across a diverse array of subjects, encompassing the Uniform Bar Exam and GRE. Furthermore, an independent study conducted by Microsoft reveals that GPT-4 outperforms the USMLE, the comprehensive medical residents' professional examination, by a significant margin \cite{J5}. Holmes et al. \cite{J6} explore the utilization of LLMs in addressing radiation oncology physics inquiries, offering insights into the scientific and medical realms. This research serves as a valuable benchmark for evaluating the performance in radiation oncology physics scenarios of LLMs. 

Unlike the studies above that used traditional assessment research, Zhuang et al. \cite{J8} introduced a novel cognitive science-based \cite{J7} methodology for evaluating LLMs. Specifically, inspired by computerized adaptive testing (CAT) in psychometrics, they proposed an adaptive testing framework for evaluating LLMs that adjusts the characteristics of test items, such as difficulty level, based on the performance of individual models. They performed fine-grained diagnosis on the latest 6 instruction-tuned LLMs (i.e. ChatGPT (OpenAI), GPT-4 (OpenAI), Bard (Google), ERNIEBot (Baidu), QianWen (Alibaba), Spark (iFlytek)) and ranked them from three aspects of Subject Knowledge, Mathematical Reasoning, and Programming. The findings demonstrate a noteworthy superiority of GPT-4 over alternative models, achieving a cognitive proficiency level comparable to that of middle-level students. Similarly, traditional evaluation approaches are also not suitable for code generation tasks. Zheng et al. \cite{J9} presented a multilingual model with 13 billion parameters for code generation, and building upon HumanEval (Python only), and they developed the HumanEval-X benchmark for evaluating multilingual models by hand-writing the solutions in C++, Java, JavaScript, and Go.

Nevertheless, a conspicuous dearth of assessment pertaining to substantial models in the realm of NLP within the domain of radiology persists. Consequently, this investigation endeavors to furnish an analytical appraisal of substantial models operating in the field of radiology. This inquiry represents the pioneering endeavor to encompass an exhaustive evaluation of large-scale language models (LLMs) within the purview of radiology, thereby serving as a catalyst for future investigations aimed at appraising the efficacy of LLMs within intricately specialized facets of medical practice.

\subsection{Large Language Model Development in Other Countries}
Many teams in other countries have conducted numerous attempts and research on LLMs. Examples include MOSS, developed by the OpenLMLab team at Fudan University, the BaiChuan series models developed by the Baidu team, the Sun and Moon model developed by SenseTime, PanGu developed by Huawei, ChatGLM-med developed by HIT, and YuLan-Chat developed by RUC. An introduction to these models will follow.
\par 
MOSS \cite{sun2023moss} is an open-source series of large language models in Chinese and English, consisting of multiple versions. The base model, moss-moon-003-base, is the foundation of all MOSS versions and is pre-trained on high-quality Chinese and English corpora, encompassing 700 billion words. To adapt MOSS to dialogue scenarios, the model moss-moon-003-sft is fine-tuned on over 1.1 million rounds of dialogue data using the base model. It possesses the capabilities of instruction-following, multi-turn dialogue comprehension, and avoidance of harmful requests. Furthermore, in addition to providing consultation and question-answering functionalities, the moss-moon-003-sft-plugin model is trained not only on over 1.1 million dialogue data but also on more than 300,000 enhanced dialogue data with plugins. It extends the capabilities of the moss-moon-003-sft model to include plugin functionalities such as using search engines, generating images from text, performing calculations, and solving equations. Both moss-moon-003-sft and moss-moon-003-sft-plugin have versions with 4-bit quantization and 8-bit quantization, making them suitable for lower resource environments.
\par 
BaiChuan, an open-source large-scale pretraining language model developed by BaiChuan Intelligence, is based on the Transformers \cite{vaswani2017attention} architecture. It has been trained on 1.2 trillion tokens and contains 7 billion parameters, supporting both Chinese and English languages. During training, a context window length of 4096 tokens was used. In actual testing, the model can also scale to over 5000 tokens. It achieved the best performance among models of the same size on standard Chinese and English language benchmarks (C-Eval \cite{huang2023ceval}/MMLU \cite{hendrycks2021measuring}).
\par 
Chat-GLM-6B is an open bilingual language model based on the General Language Model(GLM) \cite{du2022glm} framework, with 6.2 billion parameters. The model is trained for about 1 trillion tokens of Chinese and English corpus, supplemented by supervised fine-tuning, feedback bootstrap, and reinforcement learning with human feedback.\cite{ouyang2022training} It is optimized for Chinese QA and dialogue. Furthermore, the model is capable of generating answers that align with human preference.Chat-GLM-med \cite{ChatGLM-Med} is a variation of Chat-GLM-6B that is fine-tuned specifically for Chinese medical instructions. This Chinese medical instruction dataset is constructed using a medical knowledge graph and the GPT3.5 API. The fine-tuning process was performed on top of the existing Chat-GLM-6B model.
\par 
YuLan \cite{YuLan-Chat} is developed by the GSAI team at the Renmin University of China. It utilizes the LLaMA base model and is fine-tuned on a high-quality dataset of Chinese and English instructions. The dataset construction involves three stages: Open-source Instruction Deduplication, Instruction Diversification based on Topic Control, and Instruction Complexification. These stages aim to enhance the diversity of the instruction learning dataset.
\par 
There are various types of LLMs available at the current stage, and most of them are based on a base model which is a pre-trained language model following the Transformer \cite{vaswani2017attention} architecture. These models are further fine-tuned using domain-specific or high-quality data constructed on top of the base model. By leveraging high-quality, domain-specific data, a model with specialized knowledge can be derived from the base model.

\subsection{Applications in the Medical Field}
The development of LLMs could result in many potential applications in the medical field. In general, they could be used in the following four areas: clinical documentation, clinical decision support, knowledge-based medical information retrieval and generation, and medical research.

There is a large amount of clinical writing to be done by physicians and clinical professionals every day. Some can be quite laborious and time-consuming. LLMs can be possibly applied to assist in documenting patient information and symptoms \cite{lehman2023}, generating accurate and comprehensive clinical notes and test reports, and thus effectively reducing the writing load of physicians and clinical professionals. For example, an LLM summarizing the Impression from the radiology report was developed, presenting a paradigm in applications in similar domains \cite{ma2023impressiongpt}.

LLMs can also be applied to provide clinical decision support through recommending medicine usage \cite{Harskamp2023.03.25.23285475}, identifying appropriate imaging services from clinical presentations \cite{RAO2023}, or determining the cause of disease from numerous clinical notes and reports. When integrated with other modalities, like imaging, it can generate comprehensive information, assist physicians in disease diagnosis \cite{lyu2023macawllm}. In addition, from cases of patients with similar symptoms, LLMs can generate patient disease outcomes, giving a prediction of what the treatment may look like, and supporting the physicians and patients make decisions on treatment options. 

Knowledge-based application is another place where LLMs can play a big role. For example, it could be useful to have an LLM application developed to answer health-related questions from patients \cite{liu2023summary,nov2023putting}. With the training of data in specific domains, LLM applications could provide physicians and health professionals with relevant medical information from a vast amount of scientific literature, research papers, and clinical guideline, enabling quick access to up-to-date information on disease, treatments, drug interactions, and more \cite{holmes2023evaluating,liu2023radiologygpt}. Knowledge-based LLMs can help educate medical trainees and patients by answering generic or specific questions \cite{latif2023artificial}. By integrating with a patient’s medical record, LLMs can provide personalized information and explanation of drug usage, ongoing treatment, or any relevant questions patients may have.

LLMs can be of great benefit to the medical research community \cite{liu2023summary,wang2023prompt,zhang2023biomedgpt,liu2023pharmacygpt,cai2023exploring,zhong2023chatabl,wang2023review} and public health \cite{dai2023ad,guan2023cohortgpt}. For instance, the privacy of patient medical records is a big concern in clinics. The removal of identification information is mandatory before medical records used for research and results are released to the public. An LLM application help remove the identification information from medical records and could be widely utilized and beneficial to medical research \cite{liu2023deid}. Training of clinical NLP models may suffer from a lack of medical text data; augmentation of medical text data by LLMs could
provide additional samples profiting the NLP model training \cite{dai2023auggpt}. Moreover, LLMs could conduct data collection, processing, and analysis about specific diseases, providing quantified metrics and valuable insight to researchers \cite{dai2023adautogpt}.

\section{Methodology}

This section will discuss our testing methods for LLMs. We will begin by introducing the datasets MIMIC and OpenI, which we use for evaluation. Our testing approach involves employing a fixed set of prompts and parameters to assess the performance of LLMs in the field of radiology, specifically focusing on deriving impression-based performance from findings. To ensure consistency, we set several hyperparameters of the LLMs, namely the temperature to 0.9, the top\_k to 40, and the top\_p to 0.9. To evaluate the model's zero-shot and few-shot performance, we utilize zero-shot, one-shot, and five-shot examples as prompts. The experimental results and their detailed analysis are presented in the results section.

\subsection{Testing Approach}

Our testing approach involves utilizing a fixed set of prompts and parameters to evaluate the LLMs. The model's inference parameters, namely the temperature, top\_k, and top\_p, are fixed at 0.9, 40, and 0.9, respectively, to ensure consistency. We engage zero-shot, one-shot, and five-shot prompts to examine the model's zero-shot and few-shot performance. A zero-shot prompt involves presenting the model with a new task, with no prior examples provided. A one-shot prompt involves providing the model with one prior example, while a five-shot prompt provides the model with five prior examples. This variation in prompts offers a nuanced understanding of how the LLMs operate under different conditions and degrees of prior exposure.

\begin{figure}[H]
    \centering
    \includegraphics[scale=0.5]{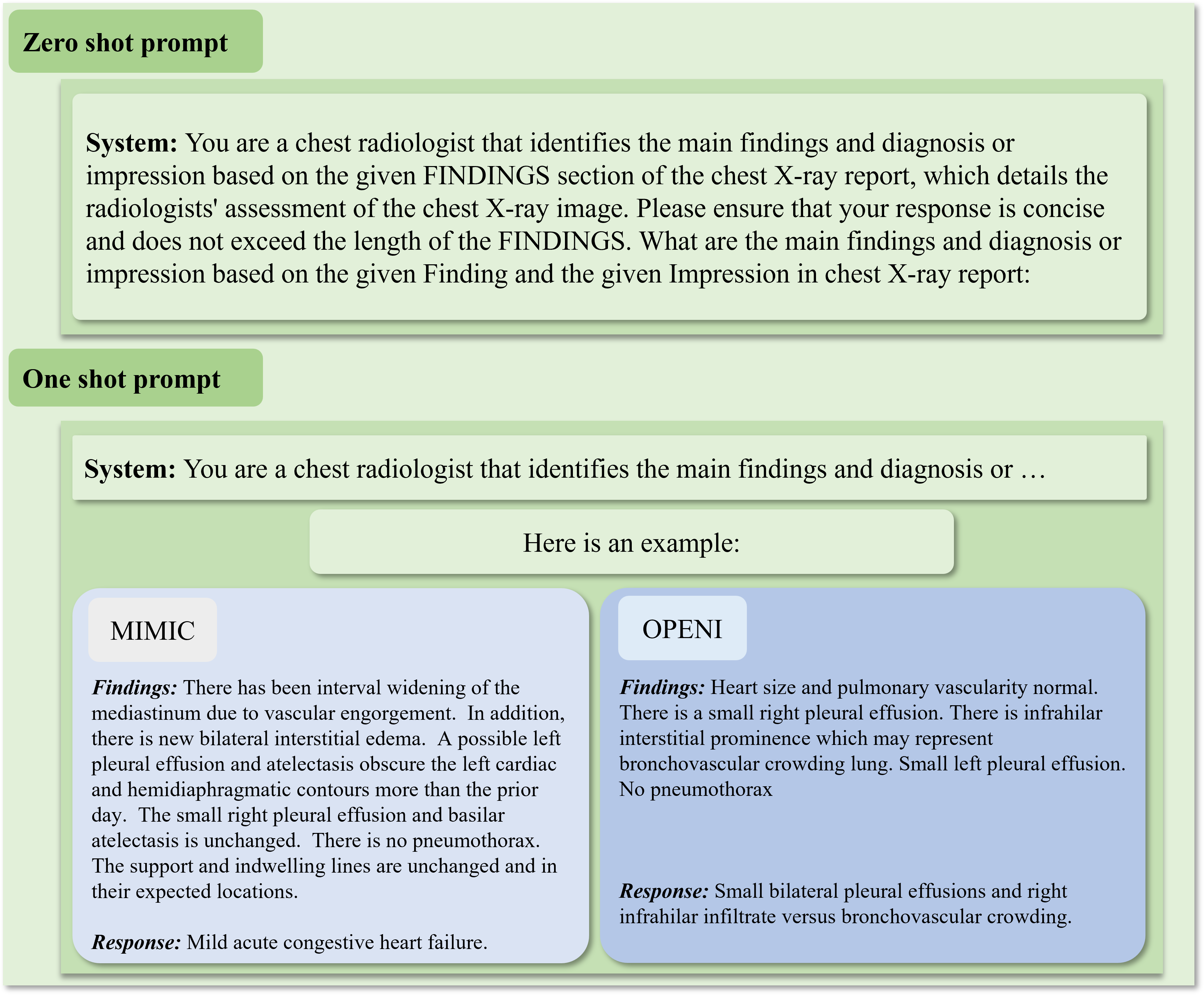}
    \caption{Zero shot prompt and one shot prompt used in the experiment.}
    \label{fig:prompt_visualization1}
\end{figure}

\begin{figure}[H]
    \centering
    \includegraphics[scale=0.5]{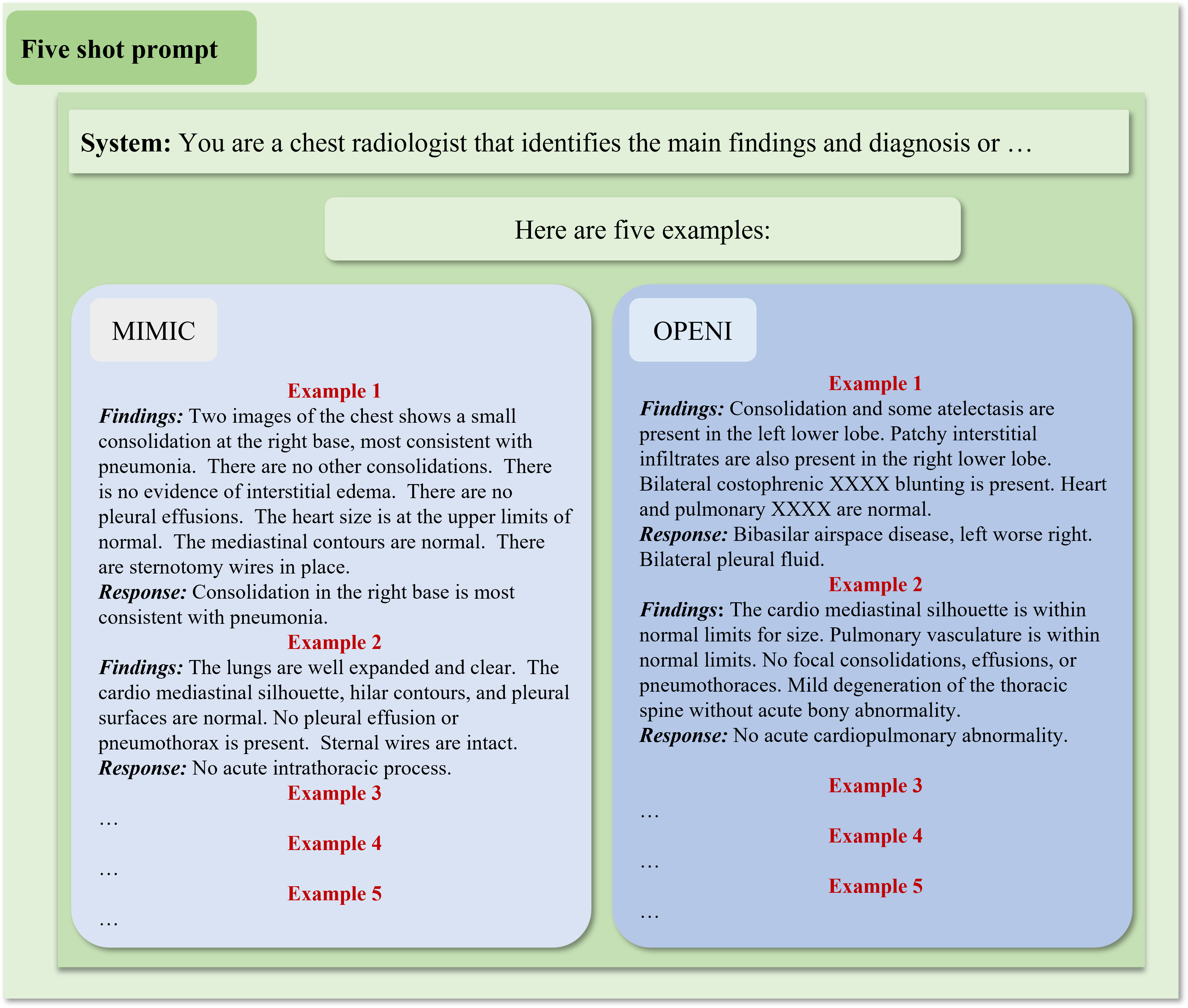}
    \caption{Five shot prompt used in the experiment.}
    \label{fig:prompt_visualization2}
\end{figure}

\subsection{Model Selection}

Considering both resource constraints and the need for uniformity in model comparison, our evaluation specifically focuses on Large Language Models (LLMs) with approximately 7 billion parameters. The choice of this parameter count is based on two primary considerations. First, models of this size strike a balance between computational efficiency and model performance. They allow for faster inference, making it feasible to thoroughly evaluate the models over the complete testing dataset in a practical timeframe. Second, this parameter count is well-represented across different types of LLMs, allowing for a broad and diverse range of models to be included in the study.

For open-source models, we procure the necessary code and model parameters directly from their official GitHub repositories. These repositories provide comprehensive documentation and community support, ensuring that the models are implemented and evaluated correctly.

For commercially available models, such as Sensenova, ChatGPT, GPT-4, PaLM2, and Anthropic Claude2, we utilize their respective Application Programming Interfaces (APIs). These APIs offer a structured and standardized way of interacting with the models, enabling us to input our pre-determined prompts and parameters and receive the model outputs in a consistent and reliable manner.

Model Summary:
\begin{enumerate}
    \item \textbf{HuatuoGPT} is a language model developed by the Shenzhen Research Institute of Big Data from the Chinese University of Hong Kong, Shenzhen. HuatuoGPT-7B is trained on the Baichuan-7B corpus, while HuatuoGPT-13B is based on Ziya-LLaMA-13B-Pretrain-v1. The advantage of HuatuoGPT is in its integration of real-world medical data and the information-rich base of ChatGPT. This allows HuatuoGPT to provide detailed diagnoses and advice in medical consultation scenarios, similar to a doctor's approach \cite{zhang2023huatuogpt}. HuatuoGPT has two versions: HuatuoGPT-7B and HuatuoGPT-13B. In our experiments, we used the HuatuoGPT-7B version.

    \item \textbf{Luotuo} is a Chinese language model exploited and maintained by the researchers Qiyuan Chen, Lulu Li, and Zihang Leng. Luotuo is fine-tuned by the LLaMA on Chinese corpus utilizing LoRA technique and does well in Chinese infering \cite{luotuo}. Luotuo has three versions: Luotuo-lora-7b-0.1, Luotuo-lora-7b-0.3, and luotuo-lora-7b-0.9. Luotuo-lora-7b-0.3 was used in the experiments.

    \item \textbf{Ziya-LLaMA} \cite{wang2022fengshenbang} denotes bilingual pre-trained language models based on LLaMA. It is a member of the open-source general large model series and is introduced by the Center for Cognitive Computing and Natural Language Research (CCNL) at the IDEA Research Institute. Ziya-LLaMA boasts remarkable versatility, demonstrating proficiency across a wide array of tasks including translation, programming, text classification, information extraction, summarization, copywriting, common sense Q\&A, and mathematical calculation. Its comprehensive training process comprises three stages: large-scale continual pre-training, multi-task supervised fine-tuning, and human feedback learning. Ziya has four version, Ziya-LLaMA-13B-v1.1, Ziya-LLaMA-13B-v1, Ziya-LLaMA-7B-Reward, and Ziya-LLaMA-13B-Pretrain-v1. In this study, we investigated the Ziya-LLaMA-13B-v1.

    \item \textbf{YuYan-Dialogue} YuYan-Dialogue \cite{li-etal-2022-easy} is a Chinese language dialogue model by fine-tuning the YuYan-11b on a large multi-turn dialogue dataset of high quality and developed by Fuxi AI lab, Netease.Inc. It is trained on a large Chinese novel dataset of high quality and has very strong conversation generation capabilities. YuYan-Dialogue has only one version that is YuYan-Dialogue. Therefore, we used it in our experiments.

    \item \textbf{BenTsao} BenTsao \cite{wang2023huatuo} is a medical language model based on LLaMA-7B model developed by SCIR Lab in Harbin Institution of Technology. It has undergone Chinese medical instruction fine-tuning and instruction tuning. They built a Chinese medical instruction dataset through the medical Knowledge graph and GPT3.5 API, based on which, they further fine-tuned the model, improving the question-and-answer effect of LLaMA in the medical field. BenTsao has four versions, LLaMA-med, LLaMA-literature, Alpaca-med, Alpaca-all-data. Here, we used the LLaMA-med (BenTsao) for comparison.

    \item \textbf{XrayGLM} Xray-GLM \cite{wang2023XrayGLM} is a vision-language model developed by Macao Polytechnic University. It is based on the VisualGLM-6B and fintuned on the translated Chinese version MIMIC-CXR, OpenI dataset. It has strong ability on chest Xray VQA. Here, we used the newest version of the Xray-GLM for comparison.

    \item \textbf{ChatGLM-Med} ChatGLM-Med \cite{ChatGLM-Med} is a language model developed by SCIR Lab in Harbin Institution of Technology. It is based on the ChatGLM-6b and has undergone Chinese medical instruction fine-tuning and instruction tuning. They built a Chinese medical instruction dataset through the medical Knowledge graph and GPT3.5 API, and on this basis, and fine-tuned the model based on the instructions of ChatGLM-6B, improving the question-and-answer effect of ChatGLM in the medical field. Here, we chose the newest version of ChatGLM-Med model for comparison.

    \item \textbf{ChatGPT/GPT4} ChatGPT and GPT4 are both highly influential large language models developed by OpenAI. The full name of ChatGPT is gpt-3.5-turbo, which is developed on the basis of gpt2 and gpt3.The training process of ChatGPT mainly refers to instructGPT \cite{ouyang2022training}, ChatGPT is an improved instructionGPT. The main difference from GPT-3  \cite{brown2020language}. is that the new addition is called RLHF (Reinforcement Learning from Human Feedback, human feedback reinforcement learning) \cite{knox2011augmenting}. This training paradigm enhances human conditioning of the model output and enables a more comprehensible ranking of the results. ChatGPT has strong language understanding ability and can handle various language expressions and queries. ChatGPT has an extensive knowledge base that can answer various frequently asked questions and provide useful information. GPT-4 is a successor to GPT-3, so it may be more capable in some ways. In our experiments, we used the ChatGPT and GPT4.

    \item \textbf{ChatGLM2/ChatGLM} ChatGLM2 is a large language model developed by Tsinghua University, developed on the basis of the ChatGLM using the GLM framework \cite{du2022glm}. ChatGLM2 has more powerful performance, which can handle longer contexts and perform more efficient reasoning with a more open protocol. What’s more, ChatGLM2 is an excellent bilingual pre-trained model \cite{zeng2022glm}. There are many versions of ChatGLM2 depending on the size of the pattern instruction set. This work mainly tested ChatGLM2-6B and ChatGLM-6B.

    \item \textbf{QiZhenGPT} QiZhenGPT \cite{ QiZhenGPT23ZJU} is a model developed by Zhejiang University. It uses the Chinese medical instruction data set constructed by QiZhen Medical Knowledge Base, and based on this, performs instruction fine-tuning on the Chinese-LLaMA-Plus-7B, CaMA-13B, and ChatGLM-6B models. QiZhenGPT has an excellent effect in Chinese medical scenarios, and it is more accurate in answering questions than ChatGLM-6B. According to different model objects fine-tuned by instructions, QizhenGPT has three types \cite{chinese-llama-alpaca}: QiZhen-Chinese-LLaMA-7B, QiZhen-ChatGLM-6B, and QiZhen-CaMA-13B. In this work, we tested mainly on QiZhen-Chinese-LLaMA-7B.

    \item \textbf{MOSS} MOSS-MOON-003 is the third version of the open-sourced plugin-augmented bilingual (i.e. Chinese and English) conversational language model MOSS, specifically from the MOSS-MOON-001 to MOSS-MOON-003, developed by the OpenLMLab from Fudan University \cite{sun2023moss}. The MOSS-MOON-003-sft is fine-tuned with supervision on approximately 1.1M multi-turn conversational data to the base model, MOSS-MOON-003-base. The advantage of MOSS-MOON-003 is it can follow bilingual multi-turn dialogues, refuse inappropriate requests and utilize different plugins due to its base model (i.e. MOSS-MOON-003-base was pre-trained on 700B English, Chinese, and code tokens), fine-tuning on multi-turn plugin-augmented conversational data, and further preference-aware training. There are 10 versions available: MOSS-MOON-003-base, MOSS-MOON-003-sft, MOSS-MOON-003-sft-plugin, MOSS-MOON-003-sft-int4, MOSS-MOON-003-sft-int8, MOSS-MOON-003-sft-plugin-int4, MOSS-MOON-003-sft-plugin-int8, MOSS-MOON-003-pm, MOSS-MOON-003, and MOSS-MOON-003-plugin. In our experiments, we used the MOSS-MOON-003-sft version.

    \item \textbf{ChatFlow} ChatFlow \cite{githubGitHubCVISZULinly} is a fully-parameterized training model developed by the Linly project team, built upon the foundations of LLaMa and Falcon and based on the TencentPretrain pre-training framework \cite{zhao2022tencentpretrain} and a large-scale Chinese scientific literature dataset \cite{li2022csl}. By utilizing both Chinese and Chinese-English parallel incremental pre-training, it transfers its language capabilities from English to Chinese. The key advantage of ChatFLow is that it addresses the issue of weaker Chinese language understanding and generation abilities found in the open-source models Falcon and LLaMa. It significantly improves the encoding and generation efficiency of Chinese texts. ChatFlow comes in two versions, namely ChatFlow-7B and ChatFlow-13B. For our experiments, we utilized the ChatFlow-7B version.

    \item \textbf{CPM-Bee} CPM-Bee \cite{githubGitHubOpenBMBCPMBee} is a large model system ecology based on OpenBMB, and it is a self-developed model of the Facing Wall team. It is a completely open source, commercially available Chinese-English bilingual basic model, and it is also the second milestone achieved through the CPM-Live training process. CPM-Bee uses the Transformer autoregressive architecture, with a parameter capacity of tens of billions, pre-training on a massive corpus of trillions of tokens, and has excellent basic capabilities. There are four versions of CPM-Bee: CPM-Bee-1B, CPM-Bee-2B, CPM-Bee-5B, CPM-Bee-10B. In this experiment, we tested the performance of CPM-Bee-5B (CPM-Bee).

    \item \textbf{PULSE} The PULSE model \cite{agrawal2022large} is a large-scale language model developed on the OpenMEDLab platform. It is based on the OpenChina LLaMA 13B model, which is further fine-tuned using approximately 4,000,000 SFT data from the medical and general domains. PULSE supports a variety of natural language processing tasks in the medical field, including health education, physician exam questions, report interpretation, medical record structuring, and simulated diagnosis and treatment. PULSE has two versions, PULSE\_7b and PULSE\_14b. In this experiment, we tested the version of PULSE\_7b.

    \item \textbf{Baichuan} Baichuan, developed by Baichuan Intelligence, is a large pre-trained model based on the Transformer architecture. The baichuan-7B model, comprising 7 billion parameters, was trained on approximately 12 trillion tokens, utilizing the same model design as LLaMa. Subsequently, they further developed the baichuan-13B model, which is even larger in size and trained on a greater amount of data \cite{baichuan2023bai}. The key advantage of the Baichuan model lies in its use of an automated learning-based data weighting strategy to adjust the data distribution during training, resulting in a language model that supports both Chinese and English. It has demonstrated robust language capabilities and logical reasoning skills across various datasets. Two versions of the Baichuan model are developed: baichuan-7B and baichuan-13B. For our experiments, we utilized the baichuan-7B version.

    \item \textbf{AtomGPT} AtomGPT \cite{AtomGPT23Atomecho}, developed by Atom Echo, is a large language model based on the model architecture of LLaMA \cite{de2023evaluation}. AtomGPT uses a large amount of Chinese and English data and codes for training, including a large number of public and non-public data sets. Developers use this method to improve model performance. AtomGPT currently has four versions: AtomGPT\_8k, AtomGPT\_14k, AtomGPT\_28k, AtomGPT\_56k. In this experiment, we chose AtomGPT\_8k for testing.


    \item \textbf{ChatYuan} ChatYuan \cite{clueai2023chatyuan} large v2 is an open-source large language model for dialogue, supports both Chinese and English languages, and in ChatGPT style. It is published by ClueAI. ChatYuan large v2 can achieve high-quality results on simple devices that allows users to operate on consumer graphics cards, PCs, and even cell phones. It got optimized for fine-tuning data, human feedback reinforcement learning, and thought chain. Also, comparing with its previous version, the model is optimized in many language abilities, like better at both Chinese and English, generating codes and so on. ChatYuan has three versions: ChatYuan-7B, ChatYuan-large-v1, ChatYuan-large-v2. In our experiments, we tested the ChatYuan-large-v2.

    \item \textbf{Bianque-2.0} Bianque \cite{chen2023bianque1} is a large model of healthcare conversations fine-tuned by a combination of directives and multiple rounds of questioning conversations. Based on BianQueCorpus, South China University of Technology chose ChatGLM-6B as the initialization model and obtained BianQue after the instruction fine-tuning training. BianQue-2.0 expands the data such as drug instruction instruction, medical encyclopedic knowledge instruction, and ChatGPT distillation instruction, which strengthens the model's suggestion and knowledge query ability. By using Chain of Questioning, the model can relate more closely to life and to improve questioning skills, which is different from most language model. It has two versions: Bianque-1.0 and Bianque-2.0. In our experiments, we tested Bianque-2.0.

    \item \textbf{AquilaChat} AquilaChat \cite{AquilaChat23BeijingAcademyofArtificialIntelligence} is a language model developed by the Beijing Academy of Artificial Intelligence. AquilaChat is an SFT model based on Aquila for fine tuning and Reinforcement learning. The AquilaChat dialogue model supports smooth text dialogue and multiple language class generation tasks. By defining extensible special instruction specifications, AquilaChat can call other models and tools, and is easy to expand its functions. AquilaChat has two versions: AquilaChat-7B and AquilaChat-33B. In our experiments, we used the AquilaChat-7B version.

    \item \textbf{Aquila} Aquila \cite{Aquila23BeijingAcademyofArtificialIntelligence} is a language model developed by the Beijing Academy of Artificial Intelligence. Aquila-7B is a basic model with 7 billion parameters. The Aquila basic model inherits the architectural design advantages of GPT-3, LLaMA, etc. in terms of technology, replaces a batch of more efficient low-level operator implementations, redesigns and implements the Chinese English bilingual tokenizer, upgrades the BMTrain parallel training method, and achieves nearly 8 times the training efficiency compared to Magtron+DeepSpeed Zero-2. Aquila has two versions: Aquila-7B and Aquila-33B. In our experiments, we used the Aquila -7B version.

    \item \textbf{Chinese-Alpaca-Plus} Chinese-Alpaca-Plus \cite{chinese-llama-alpaca} is a language model developed by Yiming Cui etc. Chinese-Alpaca-Plus is a language model based on LLaMA. Chinese-Alpaca-Plus has improved its coding efficiency and semantic understanding of Chinese by adding 20000 Chinese tags to the existing Glossary of LLaMA \cite{chinese-llama-alpaca}. Chinese-Alpaca-Plus has three versions: Chinese-Alpaca-Plus-7B, Chinese-Alpaca-Plus-13B, and Chinese-Alpaca-Plus-33B. In our experiments, we used the Chinese-Alpaca-Plus-7B version.

    \item \textbf{TigerBot} Tigerbot-7b-sft-v1 \cite{TigerBot23TigerResearch} is a language model developed by the Tigerbot Company. TigerBot-7b-sft-v1 is a large-scale language model with multiple languages and tasks. Tigerbot-7b-sft-v1 is an MVP version that has undergone 3 months of closed development and over 3000 experimental iterations.Functionally, Tigerbot-7b-sft-v1 already includes the ability to generate and understand most of the classes, specifically including several major parts: content generation, image generation, open-ended Q\&A, and long text interpretation.Tigerbot-7b-sft has two versions: Tigerbot-7b-sft-v1 and Tigerbot-7b-sft-v2. In our experiments, we used the tigerbot-7b-sft-v1 version.

    \item \textbf{XrayPULSE} XrayPULSE \cite{XrayPULSE23OpenMEDLab} is an extension of PULSE and made by OpenMEDLab. OpenMEDLab utilize MedCLIP as visual encoder and Q-former (BLIP2) following a simple linear transformation as the adapter to inject the image to PULSE. For aligning the frozen visual encoder and the LLM by the adapter, OpenMEDLab generate Chinese-version Xray-Report paired data from radiology. By extending PULSE, XrayPULSE is fine-tuned on Chinese-version Xray-Report paired datasets and aims to work as a biomedical multi-modal conversational assistant. The basic model is PULSE and we did the tests on XrayPULSE by modifying the Checkpoint file.

    \item \textbf{DoctorGLM} DoctorGLM \cite{xiong2023doctorglm} is the first chinese diagnosis large language model (released at 3rd april 2023) that developed by ShanghaiTech University \cite{xiong2023doctorglm}. It is fine-tuned on ChatGLM-6B using real-world online diagnosis dialogue. DoctorGLM has several updates and two different parameter-efficient finetune setting (p-tuning and LoRA). In our experiments, we used the DoctorGLM-5-22 p-tuning version.

    \item \textbf{Robin-7B-medical} Robin-medical (LMFlow) \cite{lmflow} is a toolkit providing a complete fine-tuning workflow for a large foundation model to support personalized training with limited computing resources. It is developed by Diao et al. from the Hong Kong University of Science and Technology. They provide a series of LoRA models based on the LLama model called Robin-medical, which are specially fine-tuned on the PubMedQA and MedMCQA datasets. The advantage of LMFlow is that it introduces an extensible and lightweight toolkit to simplify the fine-tuning and inference of general large foundation models. This allows people to fine-tune foundation models to mitigate the current status that most existing models exhibit a major deficiency in specialized-task applications. Robin-medical has 7B, 13B, 33B and 65B versions. We tested the 7B version in our experiments.

    \item \textbf{PaLM2} PaLM2 is a large language model developed by Google. PaLM 2 is a language model based on a tree structure, which makes use of the context and grammatical rules in the language to make the model's understanding of text information more refined, accurate and comprehensive. Different from traditional sequence-based models (such as GPT), PaLM2 uses some new methods that are more popular than traditional methods, such as Tree-LSTM \cite{le2015compositional}, Bert \cite{devlin2018bert}, etc. Compare to PaLM, PaLM2 excels at advanced reasoning tasks including code and math, classification and question answering, translation and multilingualism It excels at advanced reasoning tasks including code and math, classification and question answering, translation and multilingualism. It’s also being used in other state-of-the-art models, like Med-PaLM2 and Sec-PaLM. We tested the PaLM2 version in our experiments.

    \item \textbf{SenseNova} SenseNova \cite{SenseNova23SenseTime} is a large language model developed by SenseTime. Through the trinity flywheel of data, model training and deployment, it can provide various large models and capabilities such as natural language, content generation, automatic data annotation, and custom model training. Based on the previous accumulation of NLP work by SenseTime, SenseNova is still good in the domestic large language model. Based on the "SenseNova" large-scale model system, SenseTime has also developed a series of generative AI models and applications including Miahua SenseMirage, Ronin SenseAvatar, Qiongyu SenseSpace, and Gewu SenseThings. We mainly tested SenseNova in this work.

    \item \textbf{Anthropic Claude2} Claude2 is a large language model developed by Anthropic, which is characterized by helpful and trustworthy. It is developed on the basis of Claude1.3. Anthropic uses a technical framework they call Constitute AI \cite{bai2022constitutional} to achieve harmless processing of language models. Claude2 has a more powerful text processing function than GPT4, can handle larger-scale text, and has stronger context understanding ability and Chinese understanding ability. Claude is currently available in two versions, the powerful Claude, which excels at a wide range of tasks from complex dialogue and creative content generation to detailed instruction following, and the faster and more affordable Claude Instant, which also Can handle casual conversations, text analysis, summarization, and document question answering. We tested the latest version of Anthropic Claude2 for this work.

    \item \textbf{BayLing} Bayling \cite{bayling} is an instruction-following large language model equipped with advanced language alignment. It is a product from Natural Language Processing Group, Institute of Computing Technology, Chinese Academy of Science. BayLing can be effortlessly deployed on a consumer-grade GPU. It shows superior capability in English/Chinese generation, instruction following and multi-turn interaction. Bayling has three versions: BayLing-7B-v1.0, BayLing-13B-v1.0, BayLing-13B-v1.1. In our experiments, we tested BayLing-7B.

\end{enumerate}

\subsection{Uniform Testing Prompts}

For a fair and equitable comparison across different LLMs, we adopt a uniform approach in the selection and use of testing prompts. The same prompts are used across all models and conditions, regardless of whether they are zero-shot, one-shot, or five-shot scenarios.

In a zero-shot evaluation, the models are presented with a new task, with no prior examples given. For the one-shot scenario, we provide the model with one prior example. Meanwhile, in the five-shot scenario, the model is given five examples to learn from. These scenarios aim to mimic real-world usage conditions where models are given a limited number of examples and are expected to generalize from them.

\begin{figure}
    \centering
    \includegraphics[scale=0.7]{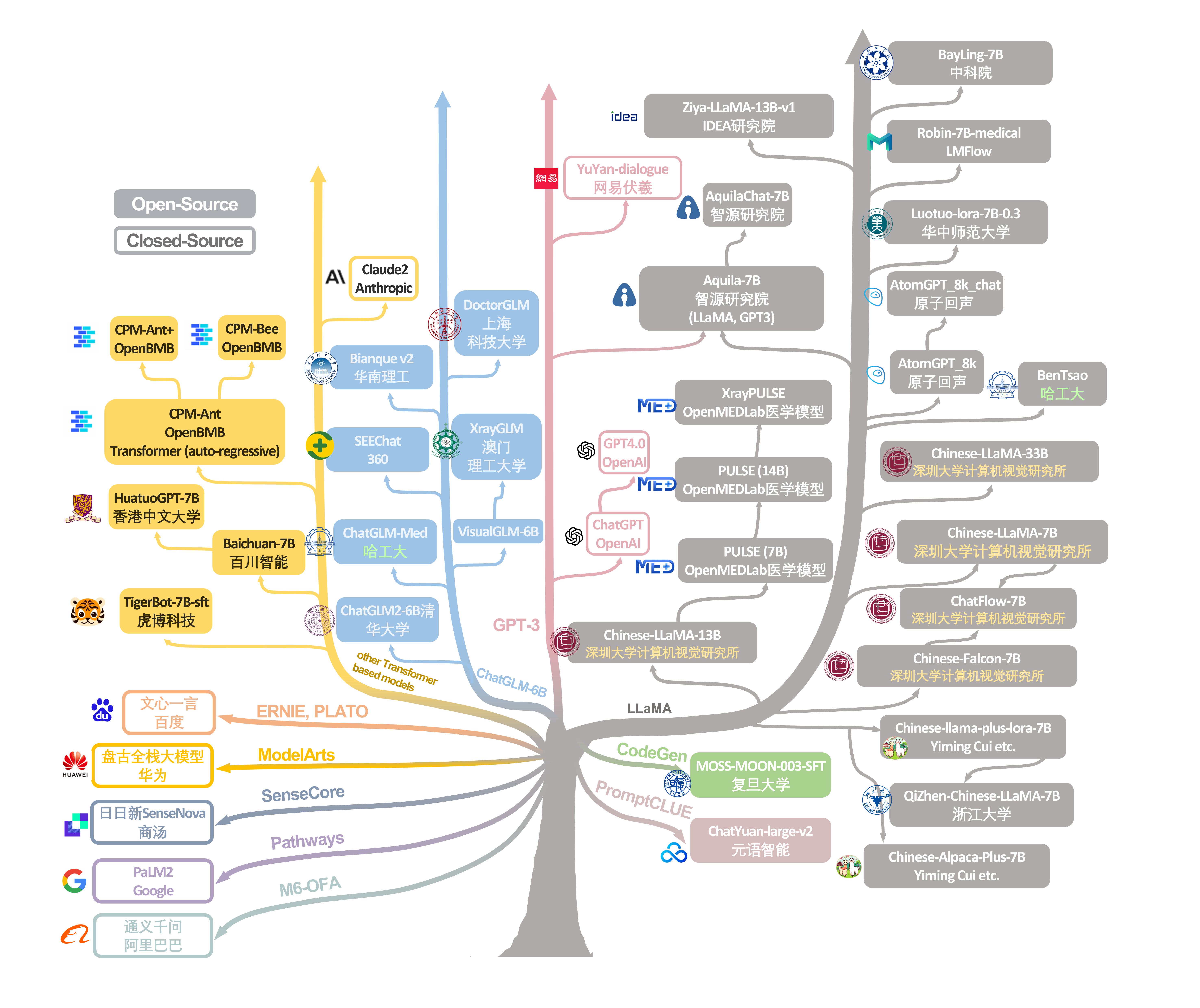}
    \caption{Major world-wide LLMs and their relationships.}
    \label{fig:model_tree}
\end{figure}

\begin{table}[ht]
\centering
\caption{Test Results for Compared LLMs}
\begin{adjustbox}{width=1\textwidth}
\small
\begin{tabular}{lccccccccccccccccccc}
\toprule
\multirow{2}{*}{Model} & \multicolumn{9}{c}{OpenI} & \multicolumn{9}{c}{MIMIC-CXR}\\
\cmidrule(lr){2-10} \cmidrule(lr){11-19}
& \multicolumn{3}{c}{zero-shot} & \multicolumn{3}{c}{one-shot} & \multicolumn{3}{c}{five-shot} & \multicolumn{3}{c}{zero-shot} & \multicolumn{3}{c}{one-shot} & \multicolumn{3}{c}{five-shot}\\
\cmidrule(lr){2-4} \cmidrule(lr){5-7} \cmidrule(lr){8-10} \cmidrule(lr){11-13} \cmidrule(lr){14-16} \cmidrule(lr){17-19}
& R-1 & R-2 & R-L & R-1 & R-2 & R-L & R-1 & R-2 & R-L & R-1 & R-2 & R-L & R-1 & R-2 & R-L & R-1 & R-2 & R-L\\
\midrule
Luotuo-lora-7B-0.3 & 0.147 & 0.0495 & 0.1262 & 0.152 & 0.0561 & 0.1356 & 0.4282 & 0.3238 & 0.4168 & 0.2366 & 0.1229 & 0.1899 & 0.0795 & 0.0132 & 0.0761 & 0.2649 & 0.1424 & 0.2491 \\
Ziya-LLaMA-13B-v1 & 0.1101 & 0.0316 & 0.0926 & 0.1502 & 0.0285 & 0.1379 & 0.2794 & 0.1694 & 0.2702 & 0.2121 & 0.0968 & 0.17 & 0.1103 & 0.0299 & 0.0995 & 0.2806 & 0.139 & 0.2512 \\
YuYan-dialogue & 0.0622 & 0.0183 & 0.0599 & 0.0702 & 0.009 & 0.0627 & 0.0981 & 0.0529 & 0.092 & 0.0819 & 0.0303 & 0.0978 & 0.0557 & 0.0123 & 0.0607 & 0.1193 & 0.0807 & 0.1231 \\
BenTsao  & 0.0804 & 0.0368 & 0.0711 & 0.0912 & 0.0201 & 0.0977 & 0.1302 & 0.0803 & 0.1231 & 0.1319 & 0.0618 & 0.1126 & 0.0604 & 0.0423 & 0.0627 & 0.1687 & 0.0807 & 0.1667 \\
XrayGLM & 0.0622 & 0.0183 & 0.0599 & 0.0808 & 0.0063 & 0.0631 & 0.0612 & 0.0215 & 0.0501 & 0.1104 & 0.0468 & 0.1211 & 0.0538 & 0.0233 & 0.0531 & 0.1533 & 0.0621 & 0.1611 \\
ChatGLM-Med & 0.0796 & 0.0317 & 0.0661 & 0.0702 & 0.025 & 0.0815 & 0.1201 & 0.0713 & 0.1488 & 0.1233 & 0.0586 & 0.1344 & 0.0702 & 0.0335 & 0.0641 & 0.1601 & 0.0806 & 0.1744 \\
ChatGPT & 0.1203 & 0.037 & 0.1052 & 0.1363 & 0.0421 & 0.1205 & 0.4262 & 0.2961 & 0.4113 & 0.2048 & 0.0996 & 0.1702 & 0.2506 & 0.1183 & 0.2052 & 0.3401 & 0.1871 & 0.2921 \\
GPT4 & 0.1171 & 0.0343 & 0.0975 & 0.1079 & 0.0328 & 0.0909 & 0.1357 & 0.0617 & 0.1176 & 0.1995 & 0.0858 & 0.1575 & 0.2114 & 0.0875 & 0.163 & 0.2029 & 0.0904 & 0.1607 \\
ChatGLM2-6B & 0.1094 & 0.0331 & 0.0909 & 0.0976 & 0.0237 & 0.0886 & 0.215 & 0.1346 & 0.2058 & 0.2042 & 0.0964 & 0.1605 & 0.0974 & 0.033 & 0.0844 & 0.247 & 0.1254 & 0.2198 \\
ChatGLM-6B & 0.125 & 0.0398 & 0.1087 & 0.1576 & 0.0511 & 0.143 & 0.2568 & 0.1451 & 0.2432 & 0.2051 & 0.1007 & 0.1694 & 0.1439 & 0.0464 & 0.1293 & 0.2766 & 0.1409 & 0.2386 \\
QiZhen-Chinese-LLaMA-7B & 0.1122 & 0.0281 & 0.0965 & 0.1033 & 0.0259 & 0.0823 & 0.224 & 0.1461 & 0.2123 & 0.2137 & 0.0877 & 0.1743 & 0.1021 & 0.041 & 0.0901 & 0.258 & 0.1254 & 0.2198 \\
MOSS-MOON-003-sft & 0.1402 & 0.0341 & 0.1241 & 0.1275 & 0.0242 & 0.1143 & 0.2206 & 0.1088 & 0.2057 & 0.2203 & 0.0914 & 0.1757 & 0.1258 & 0.0363 & 0.1062 & 0.2391 & 0.0999 & 0.1957 \\
ChatFlow-7B & 0.1048 & 0.0099 & 0.0978 & 0.0942 & 0.0165 & 0.0829 & 0.1844 & 0.0733 & 0.1722 & 0.1145 & 0.022 & 0.0933 & 0.105 & 0.0169 & 0.0857 & 0.1401 & 0.0409 & 0.1166 \\
CPM-Bee &  0.1193 & 0.0374 & 0.1048 & 0.1548 & 0.0297 & 0.1465 & 0.1582 & 0.0327 & 0.1501 & 0.2022 & 0.0938 & 0.1686 & 0.1125 & 0.0437 & 0.0993 & 0.1354 & 0.051 & 0.1184 \\
PULSE-7B & 0.1286 & 0.0413 & 0.1111 & 0.0885 & 0.0319 & 0.0726 & 0.1198 & 0.0576 & 0.1028 & 0.2559 & 0.1246 & 0.2043 & 0.1827 & 0.0796 & 0.1382 & 0.1298 & 0.0503 & 0.0955 \\
Baichuan-7B & 0.003 & 0.0009 & 0.0028 & 0.1328 & 0.0472 & 0.1172 & 0.2485 & 0.1467 & 0.2379 & 0.0057 & 0.0029 & 0.0042 & 0.1746 & 0.0804 & 0.1456 & 0.2301 & 0.1229 & 0.2032 \\
Chinese-Falcon-7B & 0.0518 & 0.0168 & 0.0416 & 0.0465 & 0.0154 & 0.0364 & 0.0378 & 0.0176 & 0.0312 & 0.1119 & 0.0557 & 0.0867 & 0.094 & 0.0448 & 0.0721 & 0.0491 & 0.0248 & 0.0381 \\
AtomGPT\_8k & 0.0287 & 0.0013 & 0.0245 & 0.0064 & 0.0001 & 0.0054 & 0.0024 & 0.0001 & 0.0021 & 0.0309 & 0.0014 & 0.0222 & 0.0035 & 0.0001 & 0.0029 & 0.0082 & 0.0001 & 0.0069 \\
ChatYuan-large-v2 & 0.0845 & 0.0223 & 0.0761 & 0.1005 & 0.0331 & 0.0903 & 0.2031 & 0.1078 & 0.1991 & 0.1353 & 0.0632 & 0.1151 & 0.1879 & 0.0793 & 0.15 & 0.0108 & 0.004 & 0.0102 \\
Bianque v2 & 0.0227 & 0.0036 & 0.0222 & 0.0271 & 0.0039 & 0.0264 & 0.0304 & 0.0072 & 0.0291 & 0.0294 & 0.0072 & 0.0281 & 0.0295 & 0.0062 & 0.0285 & 0.0225 & 0.0039 & 0.0212 \\
AquilaChat-7B & 0.0948 & 0.0279 & 0.0786 & 0.1079 & 0.0259 & 0.0896 & 0.2288 & 0.1454 & 0.2218 & 0.1885 & 0.0862 & 0.1474 & 0.1419 & 0.0408 & 0.1112 & 0.2084 & 0.0895 & 0.1781 \\
Aquila-7B & 0.0373 & 0.0071 & 0.0334 & 0.028 & 0.0057 & 0.0247 & 0.026 & 0.0087 & 0.0234 & 0.0707 & 0.022 & 0.0576 & 0.0411 & 0.011 & 0.0339 & 0.0425 & 0.0119 & 0.0353 \\
Chinese-Alpaca-Plus-7B & 0.063 & 0.0056 & 0.0507 & 0.0492 & 0.0035 & 0.0414 & 0.0552 & 0.0064 & 0.048 & 0.0898 & 0.0118 & 0.0663 & 0.0795 & 0.0113 & 0.0608 & 0.0807 & 0.0163 & 0.0642 \\
TigerBot-7B-sft & 0.064 & 0.0102 & 0.0527 & 0.1246 & 0.0284 & 0.1083 & 0.1562 & 0.0415 & 0.1401 & 0.1321 & 0.0331 & 0.1005 & 0.1452 & 0.043 & 0.1256 & 0.205 & 0.0676 & 0.1667 \\
XrayPULSE & 0.0293 & 0.0011 & 0.0239 & 0.0282 & 0.0018 & 0.0233 & 0.0263 & 0.0011 & 0.0225 & 0.0592 & 0.0061 & 0.0398 & 0.0531 & 0.005 & 0.0378 & 0.0488 & 0.006 & 0.0388 \\
DoctorGLM & 0.0996 & 0.0329 & 0.0861 & 0.1353 & 0.0463 & 0.1205 & 0.1392 & 0.0576 & 0.1328 & 0.1853 & 0.0916 & 0.153 & 0.0664 & 0.0146 & 0.0639 & 0.2116 & 0.1055 & 0.1916 \\
Robin-7B-medical & 0.0211 & 0.0043 & 0.0173 & 0.0335 & 0.0101 & 0.0285 & 0.0423 & 0.016 & 0.0377 & 0.0497 & 0.0139 & 0.0375 & 0.0323 & 0.0044 & 0.0236 & 0.0474 & 0.0132 & 0.0353 \\
PaLM2 & 0.1386 & 0.0477 & 0.1194 & 0.1557 & 0.0551 & 0.1353 & 0.1386 & 0.0492 & 0.1245 & 0.2749 & 0.1442 & 0.2281 & 0.2711 & 0.1446 & 0.2251 & 0.2397 & 0.1216 & 0.2019 \\
Sensenova & 0.0634 & 0.0106 & 0.051 & 0.0682 & 0.0121 & 0.055 & 0.0994 & 0.0214 & 0.0876 & 0.1136 & 0.0304 & 0.0809 & 0.1209 & 0.036 & 0.0889 & 0.1401 & 0.0395 & 0.0907 \\
Anthropic Claude2 & 0.2372 & 0.1259 & 0.2193 & 0.1944 & 0.0888 & 0.1713 & 0.4086 & 0.2755 & 0.3904 & 0.3177 & 0.153 & 0.256 & 0.3222 & 0.1514 & 0.2626 & 0.3116 & 0.1568 & 0.2548 \\
BayLing-7B & 0.1252 & 0.0389 & 0.1044 & 0.1268 & 0.0467 & 0.1181 & 0.4506 & 0.3452 & 0.4436 & 0.2149 & 0.107 & 0.1747 & 0.0851 & 0.0229 & 0.0825 & 0.2901 & 0.1722 & 0.2747 \\
\bottomrule
\end{tabular}
\end{adjustbox}
\end{table}

\subsection{Datasets}

Our study utilizes two comprehensive and publicly available datasets, the MIMIC-CXR and the OpenI datasets. These datasets were utilized to test the performance and efficacy of various LLMs in generating radiology text reports.

In our study, we used these datasets to evaluate the capabilities of the LLMs. We focused on the "Findings" and "Impression" sections of each report as they provide comprehensive and detailed textual information about the imaging findings and the radiologists’ interpretations. 

\subsubsection{MIMIC-CXR Dataset}

The MIMIC-CXR dataset is a substantial repository of de-identified chest radiographs (CXRs) that are complemented with their corresponding radiology reports. The dataset contains medical data from over 60,000 patients who were admitted to the Beth Israel Deaconess Medical Center between 2001 and 2012. The radiology reports in the MIMIC-CXR dataset typically consist of two sections: "Findings" and "Impression". The "Findings" section details observations from radiology images, while the "Impression" section provides summarized interpretations of these observations. 

\subsubsection{OpenI Dataset}

The OpenI dataset is another essential resource that was used in our study. It is a freely available repository that consists of radiology images paired with their respective reports. This dataset provided an independent external platform to validate the performance and generalizability of our LLMs across different data sources.

We followed an existing literature approach to randomly divide the dataset into separate segments for testing purposes. This division resulted in a subset of 2400, 292, and 576 reports for various testing scenarios. 

\section{Results}

This section presents the evaluation results of various large language models (LLMs) on two extensive datasets, OpenI and MIMIC-CXR. The performance of the models was assessed under three distinct shot settings: zero-shot, one-shot, and five-shot. Model performance was evaluated using three key metrics: Recall@1 (R-1), Recall@2 (R-2), and Recall@L (R-L).

\subsection{OpenI Dataset Results}

On the OpenI dataset, Anthropic Claude2 excelled in the zero-shot setting, achieving an R-1 score of 0.2372, an R-2 score of 0.1259, and an R-L score of 0.2193. These results notably surpassed those of other models under the same setting. In the one-shot scenario, the model achieving the highest R-1 score was BayLing-7B with 0.1268, followed closely by Luotuo-lora-7B-0.3 and Ziya-LLaMA-13B-v1 with scores of 0.152 and 0.1502, respectively. However, BayLing-7B was the standout performer in the five-shot setting, registering the highest scores across all metrics with an R-1 score of 0.4506, an R-2 score of 0.3452, and an R-L score of 0.4436.

\subsection{MIMIC-CXR Dataset Results}

The evaluation on the MIMIC-CXR dataset showed that the Anthropic Claude2 model retained its superior performance in the zero-shot setting, achieving an R-1 score of 0.3177, an R-2 score of 0.153, and an R-L score of 0.256. PaLM2 emerged as the leading model in the one-shot setting, delivering an R-1 score of 0.2711, an R-2 score of 0.1446, and an R-L score of 0.2251. In the five-shot scenario, the BayLing-7B model continued to outperform other models with the highest R-1 score of 0.2901, R-2 score of 0.1722, and R-L score of 0.2747.

However, some models like AtomGPT\_8k registered considerably lower performance across all shot settings and both datasets. For example, AtomGPT\_8k scored remarkably low in the OpenI zero-shot setting, with an R-1 score of 0.0287. It continued to score low across other shot settings and in the MIMIC-CXR dataset.

In conclusion, this evaluation underscores the significant diversity in the capabilities of different LLMs, emphasizing the need for careful model selection for specific tasks. The performance variance across different shot conditions has important implications for task-specific LLM selection in future research and applications.

\section{Discussion}

\subsection{Impact and Insights}
The present study has conducted one of the most exhaustive assessments of world-wide LLMs, focusing primarily on their utilization within the domain of radiology. The meticulous evaluation of these models, using extensive radiology report datasets and juxtaposing them with established global leading models, provides significant insights into their capabilities, limitations, and potential roles within the healthcare sector.

Our findings underscore that multiple LLMs perform comparably in interpreting radiology reports. This alignment points to their advanced natural language understanding skills and highlights their potential utility in enhancing radiology practice, where they can aid in automating radiological image interpretation, assisting in preliminary diagnosis, and thereby freeing up time for healthcare professionals. This is particularly beneficial in regions with limited access to radiologists or in healthcare scenarios where high volumes and time constraints pose significant challenges.

\subsection{Inter-model Differences and Implications}
While the performance of the world-wide models showed broad alignment, our results also spotlighted some disparities between the different models. This variance in strengths and weaknesses indicates that the choice of an LLM for a specific application should depend on the particular requirements of that task. Hence, a more profound understanding of these models, to which our study contributes, is critically essential for their effective deployment in the field.

\subsection{Implications of Evaluation Metrics}

The evaluation metric adopted in our study is Rouge Score, an N-gram-based method that inherently measures how well models conform to set answers. GPT-4, a universally recognized powerful model, did not outperform its counterpart, ChatGPT, nor did it surpass other models in the Rouge Score. This discrepancy invites a questioning of the significance of Rouge Score as a measure of radiology knowledge. The BayLing model, for instance, tended to produce succinct answers which, despite their brevity, may be of high quality and accuracy. On the contrary, GPT-4 may be more verbose and consider issues more comprehensively, showing some level of distrust in the input. The difference in results highlights the need to carefully interpret the evaluation scores, taking into account the unique characteristics of each model.

\subsection{Model Size and Performance}

Our analysis reveals that to achieve high performance in this specific task, there is no strict need for large models. Models with 7B parameters can produce impressive results, suggesting that we might be on the verge of a fourth industrial revolution driven by these more accessible, lightweight models. This prompts a reconsideration of the belief that model performance is strongly correlated with the size of the model. In fact, smaller models also demonstrated strong capabilities, raising the question of whether intelligence truly arises from the number of parameters and data accumulation.

\subsection{Multimodal LLMs: The Next Frontier}
The advent of multimodal LLMs, capable of managing multiple forms of input such as text and images, creates fascinating prospects for future research. Evaluating these models' aptitude to directly interpret radiological images, in addition to textual reports, could revolutionize radiology practice. These multimodal models could find uses in areas like disease detection and diagnosis, treatment planning, and patient monitoring.

\section{Conclusion}

In this comprehensive study, we rigorously evaluated the performance of 32 significant world-wide LLMs in the healthcare and radiology sector, comprising both global leading models such as ChatGPT, GPT-4, PaLM2, Claude2 and a robust suite of LLMs developed in other countries such as China. The overarching goal of this exploration was to benchmark these models in the context of interpreting radiology reports, enabling a nuanced understanding of their diverse capabilities, strengths, and weaknesses. Our findings affirm the competitive performance of many Chinese LLMs against their global counterparts, emphasizing their untapped potential in healthcare applications, particularly within radiology. This suggests a trajectory towards a future where these multilingual and diverse LLMs contribute to an enhanced global healthcare delivery system.

Looking ahead, our large-scale study's insights offer a compelling foundation for further exploratory research. There is immense scope for expanding these LLMs into different medical specialties and developing multimodal LLMs, the latter of which could handle complex and diverse data types to provide a more comprehensive understanding of patient health. However, as we navigate this evolving landscape of LLMs, it is imperative to give due consideration to their effective application and ethical deployment. In conclusion, our study hopes to catalyze further exploration and discussion, envisioning an era where LLMs significantly aid in healthcare provision and contribute to an enhanced standard of global patient care.

\newpage
\bibliography{LLM_refs}
\bibliographystyle{unsrt}

\newpage
\appendix

{\huge Appendix}

\bigskip

\begin{CJK*}{UTF8}{gbsn}

\begin{table}[h]
    \captionsetup{justification=justified,singlelinecheck=false}
    \caption{Ziya-LLaMA-13B-v1}
    \begin{tabular}{cc}
        \textbf{Findings} & \textbf{Impression} \\ [0.5ex]
        \begin{minipage}[t]{0.45\columnwidth}
            \begin{enumerate}
                \item Heart size within normal limits. No alveolar consolidation, no findings of pleural effusion or pulmonary edema. No pneumothorax.
                \item Minimally increased XXXX airspace opacities bilaterally, most prominent in the lung bases. Heart size is within normal limits. No pneumothorax or pleural effusion. Osseous structures are grossly intact.
                \item Stable enlarged cardiac silhouette. Persistent bilateral lower lobe airspace disease, not significantly XXXX compared to prior. No pleural effusion or pneumothorax. No acute bony abnormality.
                \item PA and lateral views the chest were obtained. Heart size is upper limits normal or mildly enlarged. The thoracic aorta is mildly tortuous. Pulmonary XXXX are within normal limits. No pneumothorax, pleural effusion, or focal air space consolidation.
                \item The lungs are clear. The heart and pulmonary XXXX are normal. The pleural spaces are clear. Mediastinal contours are normal. There is no pneumothorax.
            \end{enumerate}
        \end{minipage} &
        \begin{minipage}[t]{0.45\columnwidth}
            \begin{enumerate}
                \item No acute findings
                \item Minimally increased air space opacities bilaterally, most prominent in the lung bases. Findings are nonspecific, but may represent subsegmental atelectasis versus mild interstitial edema or an atypical infectious process.
                \item No significant change compared to prior. Bibasilar airspace disease may represent infection or mild edema.
                \item No acute cardiopulmonary disease.
                \item No acute cardiopulmonary disease
            \end{enumerate}
        \end{minipage} \\
    \end{tabular}
\end{table}

\begin{table}[h!]
    \captionsetup{justification=justified,singlelinecheck=false}
    \caption{MOSS-MOON-003-sft}
    \begin{tabular}{cc}
        \textbf{Findings} & \textbf{Impression} \\ [0.5ex]
        \begin{minipage}[t]{0.45\columnwidth}
            \begin{enumerate}
                \item Heart size within normal limits. No alveolar consolidation, no findings of pleural effusion or pulmonary edema. No pneumothorax.
                \item Minimally increased XXXX airspace opacities bilaterally, most prominent in the lung bases. Heart size is within normal limits. No pneumothorax or pleural effusion. Osseous structures are grossly intact.
                \item Stable enlarged cardiac silhouette. Persistent bilateral lower lobe airspace disease, not significantly XXXX compared to prior. No pleural effusion or pneumothorax. No acute bony abnormality.
                \item PA and lateral views the chest were obtained. Heart size is upper limits normal or mildly enlarged. The thoracic aorta is mildly tortuous. Pulmonary XXXX are within normal limits. No pneumothorax, pleural effusion, or focal air space consolidation.
                \item The lungs are clear. The heart and pulmonary XXXX are normal. The pleural spaces are clear. Mediastinal contours are normal. There is no pneumothorax.
            \end{enumerate}
        \end{minipage} &
        \begin{minipage}[t]{0.45\columnwidth}
            \begin{enumerate}
                \item No acute findings
                \item Minimally increased air space opacities bilaterally, most prominent in the lung bases. Findings are nonspecific, but may represent subsegmental atelectasis versus mild interstitial edema or an atypical infectious process.
                \item No significant change compared to prior. Bibasilar airspace disease may represent infection or mild edema.
                \item No acute cardiopulmonary disease.
                \item No acute cardiopulmonary disease
            \end{enumerate}
        \end{minipage} \\
    \end{tabular}
\end{table}

\begin{table}[h!]
    \captionsetup{justification=justified,singlelinecheck=false}
    \caption{BayLing-7B}
    \begin{tabular}{cc}
        \textbf{Findings} & \textbf{Impression} \\ [0.5ex]
        \begin{minipage}[t]{0.45\columnwidth}
            \begin{enumerate}
                \item Heart size within normal limits. No alveolar consolidation, no findings of pleural effusion or pulmonary edema. No pneumothorax.
                \item Minimally increased XXXX airspace opacities bilaterally, most prominent in the lung bases. Heart size is within normal limits. No pneumothorax or pleural effusion. Osseous structures are grossly intact.
                \item Stable enlarged cardiac silhouette. Persistent bilateral lower lobe airspace disease, not significantly XXXX compared to prior. No pleural effusion or pneumothorax. No acute bony abnormality.
                \item PA and lateral views the chest were obtained. Heart size is upper limits normal or mildly enlarged. The thoracic aorta is mildly tortuous. Pulmonary XXXX are within normal limits. No pneumothorax, pleural effusion, or focal air space consolidation.
                \item The lungs are clear. The heart and pulmonary XXXX are normal. The pleural spaces are clear. Mediastinal contours are normal. There is no pneumothorax.
            \end{enumerate}
        \end{minipage} &
        \begin{minipage}[t]{0.45\columnwidth}
            \begin{enumerate}
                \item No acute cardiopulmonary abnormality.
                \item Mild chronic obstructive pulmonary disease (COPD).
                \item Chronic obstructive pulmonary disease.
                \item No acute cardiopulmonary abnormality.
                \item No acute cardiopulmonary abnormality.
            \end{enumerate}
        \end{minipage} \\
    \end{tabular}
\end{table}

\begin{table}[h!]
    \captionsetup{justification=justified,singlelinecheck=false}
    \caption{Aquilachat-7b}
\begin{tabular}{cc}
    \textbf{Findings} & \textbf{Impression} \\ [0.5ex] 
    \begin{minipage}[t]{0.45\columnwidth}
        \begin{enumerate}
            \item Both lungs are clear and expanded. Heart and mediastinum normal.
            \item Heart size within normal limits. No alveolar consolidation, no findings of pleural effusion or pulmonary edema. No pneumothorax.
            \item Minimally increased XXXX airspace opacities bilaterally, most prominent in the lung bases. Heart size is within normal limits. No pneumothorax or pleural effusion. Osseous structures are grossly intact.
            \item Stable enlarged cardiac silhouette. Persistent bilateral lower lobe airspace disease, not significantly XXXX compared to prior. No pleural effusion or pneumothorax. No acute bony abnormality.
            \item PA and lateral views the chest were obtained. Heart size is upper limits normal or mildly enlarged. The thoracic aorta is mildly tortuous. Pulmonary XXXX are within normal limits. No pneumothorax, pleural effusion, or focal air space consolidation.
        \end{enumerate}
    \end{minipage} &
    \begin{minipage}[t]{0.45\columnwidth}
        \begin{enumerate}
            \item  No acute cardiopulmonary process.
            \item  No acute cardiopulmonary process.
            \item  Emphysematous disease with 除了一口热血之外，还有内脏器官出血
            \item  Regular chest radiographs demonstrate a stable enlarged cardiac silhouette. Persistent bilateral lower lobe airspace disease, not significantly different from prior studies. No pleural effusion or pneumothorax. No acute bony abnormalities.
            \item  No acute cardiopulmonary process.
        \end{enumerate}
    \end{minipage} \\
\end{tabular}
\end{table}

\begin{table}[h!]
    \captionsetup{justification=justified,singlelinecheck=false}
    \caption{Anthropic Claude 2}
\begin{tabular}{cc}
    \textbf{Findings} & \textbf{Impression} \\ [0.5ex] 
    \begin{minipage}[t]{0.45\columnwidth}
        \begin{enumerate}
            \item Both lungs are clear and expanded. Heart and mediastinum normal.
            \item Heart size within normal limits. No alveolar consolidation, no findings of pleural effusion or pulmonary edema. No pneumothorax.
            \item Minimally increased XXXX airspace opacities bilaterally, most prominent in the lung bases. Heart size is within normal limits. No pneumothorax or pleural effusion. Osseous structures are grossly intact.
            \item Stable enlarged cardiac silhouette. Persistent bilateral lower lobe airspace disease, not significantly XXXX compared to prior. No pleural effusion or pneumothorax. No acute bony abnormality.
            \item PA and lateral views the chest were obtained. Heart size is upper limits normal or mildly enlarged. The thoracic aorta is mildly tortuous. Pulmonary XXXX are within normal limits. No pneumothorax, pleural effusion, or focal air space consolidation.
        \end{enumerate}
    \end{minipage} &
    \begin{minipage}[t]{0.45\columnwidth}
        \begin{enumerate}
            \item  Impression: No acute cardiopulmonary abnormality.            \item  Impression: No acute cardiopulmonary abnormality.
            \item  Impression: Bilateral basilar airspace opacities most prominent in the lung bases. No pneumothorax or pleural effusion.
            \item  Impression: Stable cardiomegaly. Persistent bibasilar airspace disease without significant change.
            \item  Based on the given findings, the impression is: Mildly enlarged cardiac silhouette. No acute cardiopulmonary abnormality.
        \end{enumerate}
    \end{minipage} \\
\end{tabular}
\end{table}

\begin{table}[h!]
    \captionsetup{justification=justified,singlelinecheck=false}
    \caption{ChatGPT}
\begin{tabular}{cc}
    \textbf{Findings} & \textbf{Impression} \\ [0.5ex] 
    \begin{minipage}[t]{0.45\columnwidth}
        \begin{enumerate}
            \item Both lungs are clear and expanded. Heart and mediastinum normal.
            \item Heart size within normal limits. No alveolar consolidation, no findings of pleural effusion or pulmonary edema. No pneumothorax.
            \item Minimally increased XXXX airspace opacities bilaterally, most prominent in the lung bases. Heart size is within normal limits. No pneumothorax or pleural effusion. Osseous structures are grossly intact.
            \item Stable enlarged cardiac silhouette. Persistent bilateral lower lobe airspace disease, not significantly XXXX compared to prior. No pleural effusion or pneumothorax. No acute bony abnormality.
            \item PA and lateral views the chest were obtained. Heart size is upper limits normal or mildly enlarged. The thoracic aorta is mildly tortuous. Pulmonary XXXX are within normal limits. No pneumothorax, pleural effusion, or focal air space consolidation.
        \end{enumerate}
    \end{minipage} &
    \begin{minipage}[t]{0.45\columnwidth}
        \begin{enumerate}
            \item  No acute cardiopulmonary abnormality.
            \item  No acute cardiopulmonary abnormality.
            \item  Bilateral basal airspace opacities, no pneumothorax or pleural effusion.
            \item  Stable enlarged cardiac silhouette. Persistent bilateral lower lobe airspace disease, not significantly changed compared to prior. No pleural effusion or pneumothorax. No acute bony abnormality.
            \item  Main findings: Mildly enlarged heart size, mildly tortuous thoracic aorta. No pneumothorax, pleural effusion, or focal air space consolidation.

Diagnosis or impression: Mildly enlarged heart, mildly tortuous thoracic aorta without acute cardiopulmonary abnormalities.

        \end{enumerate}
    \end{minipage} \\
\end{tabular}
\end{table}

\begin{table}[h!]
    \captionsetup{justification=justified,singlelinecheck=false}
    \caption{GPT-4}
\begin{tabular}{cc}
    \textbf{Findings} & \textbf{Impression} \\ [0.5ex] 
    \begin{minipage}[t]{0.45\columnwidth}
        \begin{enumerate}
            \item Both lungs are clear and expanded. Heart and mediastinum normal.
            \item Heart size within normal limits. No alveolar consolidation, no findings of pleural effusion or pulmonary edema. No pneumothorax.
            \item Minimally increased XXXX airspace opacities bilaterally, most prominent in the lung bases. Heart size is within normal limits. No pneumothorax or pleural effusion. Osseous structures are grossly intact.
            \item Stable enlarged cardiac silhouette. Persistent bilateral lower lobe airspace disease, not significantly XXXX compared to prior. No pleural effusion or pneumothorax. No acute bony abnormality.
            \item PA and lateral views the chest were obtained. Heart size is upper limits normal or mildly enlarged. The thoracic aorta is mildly tortuous. Pulmonary XXXX are within normal limits. No pneumothorax, pleural effusion, or focal air space consolidation.
        \end{enumerate}
    \end{minipage} &
    \begin{minipage}[t]{0.45\columnwidth}
        \begin{enumerate}
            \item  Main findings are acromioclavicular arthritis. The diagnosis or impression is no acute preoperative findings.
            \item  Main findings are acromioclavicular arthritis with no acute preoperative findings. 
            \item  The main findings are minimally increased bilateral airspace opacities, most prominent in the lung bases. There is no evidence of pneumothorax or pleural effusion. 
            \item  The main findings are an enlarged cardiac silhouette, persistent bilateral lower lobe airspace disease, and no signs of pleural effusion, pneumothorax, or acute bony abnormalities. The diagnosis or impression is stable cardiac enlargement and persistent bilateral lower lobe airspace disease.
            \item The main findings are a mildly enlarged heart size, a mildly tortuous thoracic aorta, and normal pulmonary markings. There is no evidence of pneumothorax, pleural effusion, or focal air space consolidation. The impression or diagnosis would be mild cardiomegaly and a mildly tortuous thoracic aorta with no acute cardiopulmonary abnormalities.
        \end{enumerate}
    \end{minipage} \\
\end{tabular}
\end{table}

\begin{table}[h!]
    \captionsetup{justification=justified,singlelinecheck=false}
    \caption{ChatGLM2-6B}
\begin{tabular}{cc}
    \textbf{Findings} & \textbf{Impression} \\ [0.5ex] 
    \begin{minipage}[t]{0.45\columnwidth}
        \begin{enumerate}
            \item Both lungs are clear and expanded. Heart and mediastinum normal.
            \item Heart size within normal limits. No alveolar consolidation, no findings of pleural effusion or pulmonary edema. No pneumothorax.
            \item Minimally increased XXXX airspace opacities bilaterally, most prominent in the lung bases. Heart size is within normal limits. No pneumothorax or pleural effusion. Osseous structures are grossly intact.
            \item Stable enlarged cardiac silhouette. Persistent bilateral lower lobe airspace disease, not significantly XXXX compared to prior. No pleural effusion or pneumothorax. No acute bony abnormality.
            \item PA and lateral views the chest were obtained. Heart size is upper limits normal or mildly enlarged. The thoracic aorta is mildly tortuous. Pulmonary XXXX are within normal limits. No pneumothorax, pleural effusion, or focal air space consolidation.
        \end{enumerate}
    \end{minipage} &
    \begin{minipage}[t]{0.45\columnwidth}
        \begin{enumerate}
            \item No abnormal findings.Based on the given findings and impressions, the main findings and diagnosis or impression based on the given chest X-ray report are: No acute preoperative findings. No acute cardiopulmonary abnormality. Chest. Resolving pulmonary interstitial edema and pulmonary venous hypertension. Bibasilar airspace disease, left worse right. Bilateral pleural fluid.
            \item  No abnormalities.Based on the given findings and impressions, the main findings and diagnosis or impression based on the given chest X-ray report are: No acute preoperative findings. No acute cardiopulmonary abnormality. Chest. Resolving pulmonary interstitial edema and pulmonary venous hypertension. Bibasilar airspace disease, left worse right. Bilateral pleural fluid. No abnormalities.
            \item Pulmonary airspace opacity increased.
            \item  Enlarged cardiac silhouette with persistent lower lobe airspace disease. No acute bony abnormality.
            \item No abnormalities noted.

 Based on the given findings and impressions, the main findings and diagnosis or impression based on the given chest X-ray report are:

1.  No acute preoperative findings.
2.  No acute cardiopulmonary abnormality.
3.  Chest. Resolving pulmonary interstitial edema and pulmonary venous hypertension.
4.  Bibasilar airspace disease, left worse right. Bilateral pleural fluid.
5.  No abnormalities noted.

        \end{enumerate}
    \end{minipage} \\
\end{tabular}
\end{table}

\begin{table}[h!]
    \captionsetup{justification=justified,singlelinecheck=false}
    \caption{PaLM2}
\begin{tabular}{cc}
    \textbf{Findings} & \textbf{Impression} \\ [0.5ex] 
    \begin{minipage}[t]{0.45\columnwidth}
        \begin{enumerate}
            \item Both lungs are clear and expanded. Heart and mediastinum normal.
            \item Heart size within normal limits. No alveolar consolidation, no findings of pleural effusion or pulmonary edema. No pneumothorax.
            \item Minimally increased XXXX airspace opacities bilaterally, most prominent in the lung bases. Heart size is within normal limits. No pneumothorax or pleural effusion. Osseous structures are grossly intact.
            \item Stable enlarged cardiac silhouette. Persistent bilateral lower lobe airspace disease, not significantly XXXX compared to prior. No pleural effusion or pneumothorax. No acute bony abnormality.
            \item PA and lateral views the chest were obtained. Heart size is upper limits normal or mildly enlarged. The thoracic aorta is mildly tortuous. Pulmonary XXXX are within normal limits. No pneumothorax, pleural effusion, or focal air space consolidation.
        \end{enumerate}
    \end{minipage} &
    \begin{minipage}[t]{0.45\columnwidth}
        \begin{enumerate}
            \item The lungs are clear and expanded. Heart and mediastinum normal.
            \item  The cardiomediastinal silhouette is within normal limits. The lungs are clear. There is no evidence of focal consolidation, pleural effusion, or pneumothorax.
            \item Minimally increased airspace opacities bilaterally, most prominent in the lung bases. Heart size is within normal limits. No pneumothorax or pleural effusion. Osseous structures are grossly intact.
            \item "The main findings are:

- Enlarged cardiac silhouette
- Bilateral lower lobe airspace disease
- No pleural effusion or pneumothorax

The diagnosis or impression is:

- Stable enlarged cardiac silhouette
- Persistent bilateral lower lobe airspace disease, not significantly worse compared to prior."

            \item The cardiomediastinal silhouette is within normal limits. The lungs are clear. No focal consolidations, effusions, or pneumothoraces.
        \end{enumerate}
    \end{minipage} \\
\end{tabular}
\end{table}

\begin{table}[h!]
    \captionsetup{justification=justified,singlelinecheck=false}
    \caption{SenseNova}
\begin{tabular}{cc}
    \textbf{Findings} & \textbf{Impression} \\ [0.5ex] 
    \begin{minipage}[t]{0.45\columnwidth}
        \begin{enumerate}
            \item Both lungs are clear and expanded. Heart and mediastinum normal.
            \item Heart size within normal limits. No alveolar consolidation, no findings of pleural effusion or pulmonary edema. No pneumothorax.
            \item Minimally increased XXXX airspace opacities bilaterally, most prominent in the lung bases. Heart size is within normal limits. No pneumothorax or pleural effusion. Osseous structures are grossly intact.
            \item Stable enlarged cardiac silhouette. Persistent bilateral lower lobe airspace disease, not significantly XXXX compared to prior. No pleural effusion or pneumothorax. No acute bony abnormality.
            \item PA and lateral views the chest were obtained. Heart size is upper limits normal or mildly enlarged. The thoracic aorta is mildly tortuous. Pulmonary XXXX are within normal limits. No pneumothorax, pleural effusion, or focal air space consolidation.
        \end{enumerate}
    \end{minipage} &
    \begin{minipage}[t]{0.45\columnwidth}
        \begin{enumerate}
            \item According to the provided information, there are no significant findings identified by the radiologist. The text states "both lungs are clear and expanded" and the heart and mediastinum appear normal. Based on these observations, it appears that this patient has a negative chest x-ray result.
            \item  Impression: No significant abnormalities seen.
            \item"Based on the provided text and impression, here is my interpretation of the primary findings and diagnosis/impression:

Finding: Minimally increased reticulonodular densities bilaterally, with mild bronchiectasis.
Impression: Mild chronic obstructive pulmonary disease (COPD) with bronchiectasis.

Please note that this interpretation is only intended to provide an example answer format. It should be used as guidance rather than definitively diagnosing any specific condition. If you require medical advice or treatment, please consult a physician."
            \item "根据提供的Finding和Impression，主要发现和诊断或印象如下：

Main Findings: Stable enlarged cardiac silhouette. Persistent bilateral lower lobe airspace disease, not significantly changed compared to prior. No pleural effusion or pneumothorax. No acute bony abnormality.
Diagnosis or Impression: Bilateral bronchiectasis with stable heart failure."
            \item Impression: No significant changes noted.

        \end{enumerate}
    \end{minipage} \\
\end{tabular}
\end{table}

\end{CJK*}

\end{document}